\documentclass[]{fairmeta}

\usepackage{amsmath,amssymb,amsfonts,mathtools}
\usepackage{xspace}
\usepackage{float}
\usepackage{makecell}
\usepackage{multirow}

\microtypesetup{expansion=false}

\newcommand{\tablebodyfont}{%
  \small\rmfamily
  \renewcommand{\textbf}[1]{{\rmfamily\bfseries ##1}}%
}
\newcommand{\tablehead}[1]{{\sffamily\bfseries #1}}

\title{Surprise Forcing: What to Remember, When to Skip in Long Video Generation}

\author[1]{Shuwei Shi}
\author[1]{Zhen Li}
\author[1]{Muyao Niu}
\author[1]{Chuanhao Li}
\author[1]{Bo Zheng}
\author[1,\dagger]{Kaipeng Zhang}
\author[2,\dagger]{Yinqiang Zheng}

\affiliation[1]{Alaya Lab}
\affiliation[2]{The University of Tokyo}

\abstract{Streaming autoregressive diffusion makes minute-scale video synthesis practical, but its bounded context and fixed denoising schedule allocate resources uniformly across a highly non-stationary sequence. A rolling key--value cache forgets distant visual evidence even when that evidence remains important, while every generated chunk receives the same number of denoising passes irrespective of its actual difficulty. We introduce \textbf{Surprise Forcing}, a training-free framework that treats both limitations as online resource-allocation problems. A \textbf{Surprise-Gated Memory Bank} summarizes evicted frames with value-token descriptors, evaluates them using complementary global-deviation and nearest-neighbor novelty signals, and regulates admission through a feedback-controlled budget in normalized score space. Priority-based replacement and relevance-aware routing then keep the external memory compact and useful. In parallel, \textbf{Surprise-Aware Denoising} estimates chunk difficulty from the maximum adjacent-frame cosine distance after the first denoising pass and uses a local percentile scheduler to skip intermediate steps for comparatively easy chunks. Experiments on VBench, VBench-Long, and VBench-2.0 show that the proposed allocation strategy improves long-horizon consistency and visual quality while retaining real-time streaming throughput.}

\date{July 20, 2026}

\begin{document}

\maketitle
\begingroup
\renewcommand{\thefootnote}{\fnsymbol{footnote}}
\footnotetext[2]{Corresponding authors: kaipeng.zhang@shanda.com; yqzheng@ai.u-tokyo.ac.jp}
\endgroup

\begin{figure}[H]
    \centering
    \includegraphics[width=0.82\linewidth]{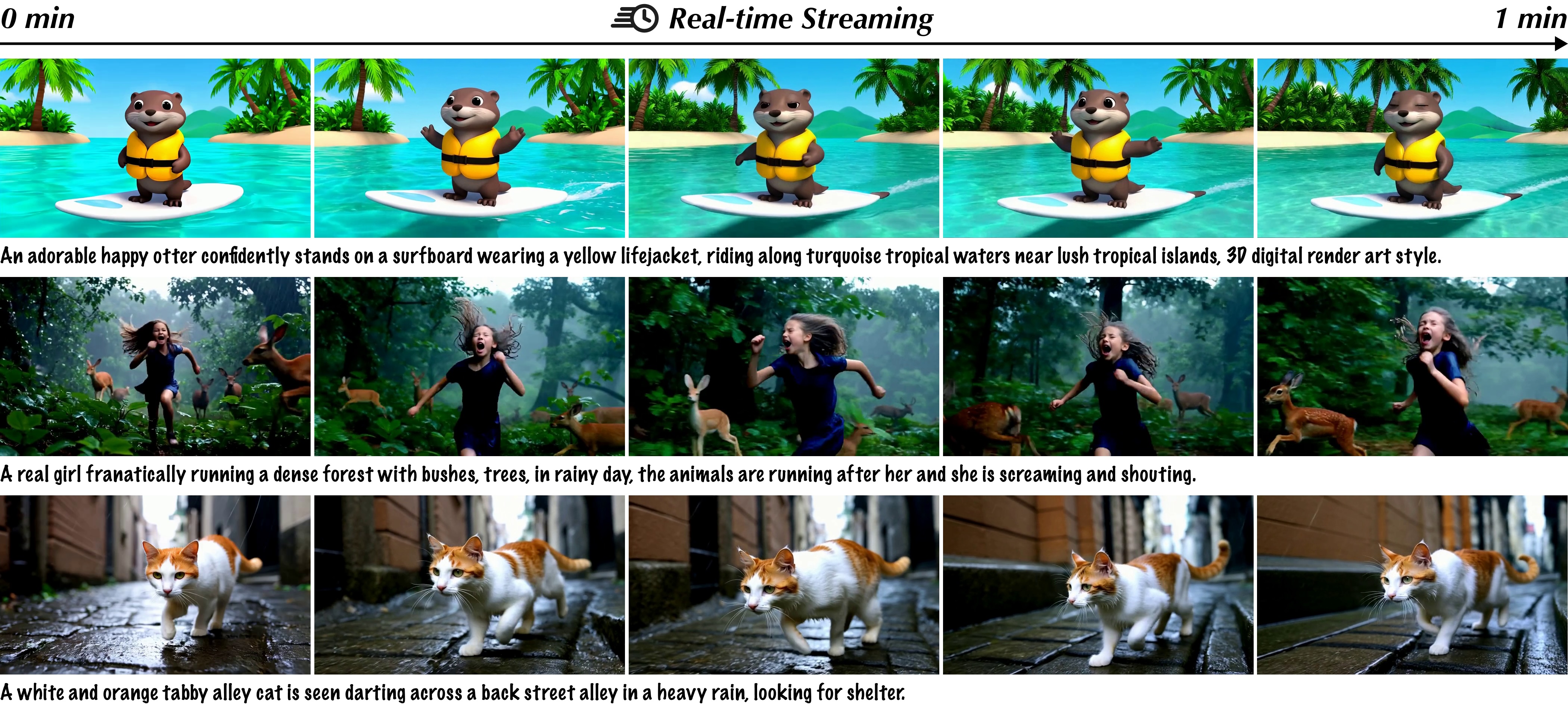}
    \caption{Samples generated by Surprise Forcing. Our method can perform real-time long video generation while maintaining good temporal consistency.}
    \label{fig:teaser}
\end{figure}

\section{Introduction}
\label{sec:intro}
Recent text-to-video diffusion models have substantially improved fidelity, motion, and prompt adherence~\cite{ho2020denoising, song2020denoising, wan2025wan, wu2025hunyuanvideo, yang2024cogvideox, tan2025mimir}. Extending these models from short clips to sustained generation, however, changes the nature of the inference problem. A streaming autoregressive (AR) system must repeatedly decide how to use a finite context window and a finite compute budget while its visual state evolves over hundreds of frames. Existing few-step AR pipelines~\cite{huang2025self, cui2025self, yang2025longlive, liu2025rolling, huang2025memory} make this regime feasible through causal chunk generation, rolling key--value (KV) caches, and distribution-matching distillation~\cite{yin2024one, yin2024improved}; they do not, by themselves, determine which historical evidence deserves continued access or which chunks deserve additional refinement.

The first issue is \textbf{selective forgetting}. A rolling cache bounds memory by discarding old states in temporal order. This policy is inexpensive, but temporal age is only weakly related to future usefulness. An early view may establish subject identity, clothing, geometry, or scene layout and remain relevant much later, whereas a recent frame may add little beyond the near-local context. Once informative history is evicted, identity drift and scene inconsistency become difficult to correct.

The second issue is \textbf{uniform computation}. A fixed denoising schedule spends the same number of transformer evaluations on every chunk. Long videos are not uniformly difficult~\cite{qiu2026histream}: nearly static continuations often stabilize quickly, while abrupt motion, object entry, viewpoint change, and scene transition require more iterative correction. Uniform refinement therefore over-computes easy regions and limits interactive throughput.

We recast both problems around a common principle: the model's own intermediate representations contain an online signal of \emph{surprise}. For memory management, surprise measures how poorly a newly evicted frame is represented by the current bank. For denoising, surprise measures how much temporal variation remains visible after the first refinement step. These signals are inexpensive, causal, and available without retraining.

Based on this principle, we introduce \textbf{Surprise Forcing}. Its first component, the \textbf{Surprise-Gated Memory Bank}, stores selected frames outside the rolling cache. A frame descriptor derived from value tokens is scored by both aggregate deviation from the bank and nearest-neighbor novelty. An adaptive controller regulates the admission rate in normalized score space; a composite priority rule decides replacement; and dynamic routing retrieves only the most relevant global context. Its second component, \textbf{Surprise-Aware Denoising}, predicts chunk difficulty from adjacent-frame latent change after one denoising step and assigns either a reduced or full schedule using a sliding-window percentile rule.

We formulate Budget-Norm Gating as a fully online controller with explicit running statistics, feedback updates, and cold-start behavior. We evaluate long-range preservation through VBench-2.0 measurements of identity, clothing, instance, and viewpoint consistency, together with controlled studies of the surprise formulation, memory capacity, routing, admission budget, eviction policy, and denoising predictor.

Our main contributions are:
\begin{itemize}
    \item a unified, training-free view of long-video inference as selective allocation of historical context and denoising computation;
    \item a surprise-gated external memory with feedback-controlled admission, utility-aware eviction, and query-dependent retrieval;
    \item a self-referential difficulty estimator that adapts denoising depth at the chunk level without an auxiliary network; and
    \item a comprehensive empirical evaluation showing how each design choice affects consistency, quality, speed, and memory use over long rollouts.
\end{itemize}

\section{Related Work}

\subsection{Autoregressive Long Video Generation}
Bidirectional video diffusion offers strong global modeling over short clips, but its cost grows rapidly with temporal extent. Training-free approaches such as FreeNoise~\cite{qiu2024freenoise}, FreeLong~\cite{lu2025freelong}, FreeLong++~\cite{lu2025freelong++}, FIFO-Diffusion~\cite{kim2024fifo}, and rolling diffusion~\cite{ruhe2024rolling} extend the sampling horizon by modifying noise schedules or temporal processing. Diffusion Forcing~\cite{chen2024diffusion} and HistoryGuidance~\cite{song2025history} instead connect diffusion with causal prediction, a direction also reflected in systems such as SkyReels-V2~\cite{chen2025skyreels} and Magi-1~\cite{teng2025magi1}.

A second line explicitly targets streaming AR generation. StreamingT2V~\cite{henschel2025streamingt2v} blends historical information, CausVid~\cite{yin2025slow} distills a bidirectional teacher into a causal student, StreamDiT~\cite{kodaira2026streamdit} combines distillation with a moving buffer, Rolling Forcing~\cite{liu2025rolling} develops real-time diffusion-forcing rollouts, and Self Forcing~\cite{huang2025self} reduces train--test mismatch through self-rollout. Self-Forcing++~\cite{cui2025self} and LongLive~\cite{yang2025longlive} further extend the usable horizon. Surprise Forcing is complementary to these training strategies: it operates only at inference time and reallocates the memory and denoising budgets of an already trained streaming generator.

\subsection{Context Management and Explicit Memory}
Bounded local attention is the standard mechanism for controlling streaming cost. Attention sinks, first analyzed for language-model streaming~\cite{xiao2024efficient}, preserve a few initial anchors, but most intermediate history is still forgotten. Deep Forcing~\cite{yi2025deep} retains high-participation tokens through sink design and compression inside the active window. By contrast, our method moves selected evicted frames into a compact external store and retrieves them at frame granularity.

Related approaches compress or package context before reuse. LoViC~\cite{jiang2026lovic}, FramePack-style conditioning~\cite{zhang2025framepack}, learned memory encoders~\cite{zhang2025pretraining}, and pack-and-force strategies~\cite{wu2025pack} trade representation detail for a longer effective horizon. Explicit retrieval has also become important in long-horizon generation. The general idea resembles retrieval-augmented generation~\cite{lewis2020retrieval}, but video memory must preserve causal visual states under strict latency constraints. WorldMem~\cite{xiao2026worldmem}, Context-as-Memory~\cite{yu2025context}, VMem~\cite{li2025vmem}, Memory Forcing~\cite{huang2025memory}, and MemFlow~\cite{ji2025memflow} learn or design retrieval mechanisms for world simulation and long narratives. Surprise Forcing differs by using a lightweight, training-free write controller and by jointly considering admission, replacement, and routing.

\subsection{Efficient Video Diffusion}
Few-step synthesis is commonly obtained through distribution matching and related distillation objectives~\cite{yin2024one, yin2024improved}; CausVid~\cite{yin2025slow} and StreamDiT~\cite{kodaira2026streamdit} apply such ideas to streaming video. Training-free caching, exemplified by AdaCache~\cite{kahatapitiya2025adaptive}, reuses intermediate transformer computation and modulates reuse according to motion. Sparse VideoGen~\cite{xi2025sparse}, Radial Attention~\cite{li2026radial}, and Mixture-of-Contexts~\cite{cai2025mixture} reduce token or attention cost through architectural sparsity. Our focus is different: the backbone and weights are unchanged, and compute is assigned at the chunk level after observing a one-step estimate of temporal difficulty.

\section{Method}
Our framework builds on a distilled streaming generator and adds two inference-time controllers. We first summarize the underlying AR formulation, then describe how surprise governs external memory and denoising depth.

\subsection{Preliminaries}
\label{sec:preliminaries}
Let a long video be generated as a sequence of fixed-length latent chunks. Conditioned on text $c$, the causal factorization is

\begin{equation}
p_\theta(\mathbf{x}_{1:N}\mid c)
=
\prod_{n=1}^{N} p_\theta(\mathbf{x}_n \mid \mathbf{x}_{<n}, c),
\label{eq:ar_factor}
\end{equation}

where $\mathbf{x}_n$ is the $n$-th chunk and the preceding chunks are represented through a bounded rolling KV cache.

The base generator follows Distribution Matching Distillation (DMD), which aligns a causal student with a short-horizon bidirectional teacher. Let $t$ be a sampled diffusion time, $\mathbf{z}\sim\mathcal{N}(0,\mathbf{I})$ the generator noise, $G_\theta(\mathbf{z})$ the student output, and $\Phi(\cdot,t)$ the forward noising operator. Denote the student and teacher distributions at time $t$ by $p^S_{\theta,t}$ and $p^T_t$, with corresponding scores $s^S_\theta$ and $s^T$. The objective is

\begin{equation}
\mathcal{L}_{\mathrm{DMD}}
=
\mathbb{E}_{t}\!\left[\mathrm{KL}\!\left(p^{S}_{\theta,t}\,\|\,p^{T}_{t}\right)\right],
\label{eq:dmd_loss}
\end{equation}

with the score-form approximation

\begin{equation}
\begin{split}
\nabla_{\theta}\mathcal{L}_{\mathrm{DMD}}
\approx
-\mathbb{E}_{t,\mathbf{z}}\!\left[
\int
\Big(
s^{T}\!\big(\Phi(G_{\theta}(\mathbf{z}),t),t\big)
-
s^{S}_{\theta}\!\big(\Phi(G_{\theta}(\mathbf{z}),t),t\big)
\Big)\,
\frac{\partial G_{\theta}(\mathbf{z})}{\partial\theta}\,
d\mathbf{z}
\right].
\end{split}
\label{eq:dmd_grad}
\end{equation}

Because the teacher covers only a short temporal horizon, training samples windows from a longer self-rollout. For a rollout of $N$ chunks, a contiguous length-$K$ window $W_i:=\mathbf{x}_{i:i+K-1}$ is sampled uniformly and re-noised before evaluating

\begin{equation}
\mathcal{L}^{\mathrm{ext}}_{\mathrm{DMD}}
=
\mathbb{E}_{t}\,
\mathbb{E}_{i\sim\mathrm{Unif}\{1,\ldots,N-K+1\}}
\left[\mathrm{KL}\!\left(p^{S}_{\theta,t}(W_i)\,\|\,p^{T}_{t}(W_i)\right)\right].
\label{eq:ext_dmd}
\end{equation}

\subsection{Overview}
\label{sec:overview}
\begin{figure}[t]
    \centering
    \includegraphics[width=\linewidth]{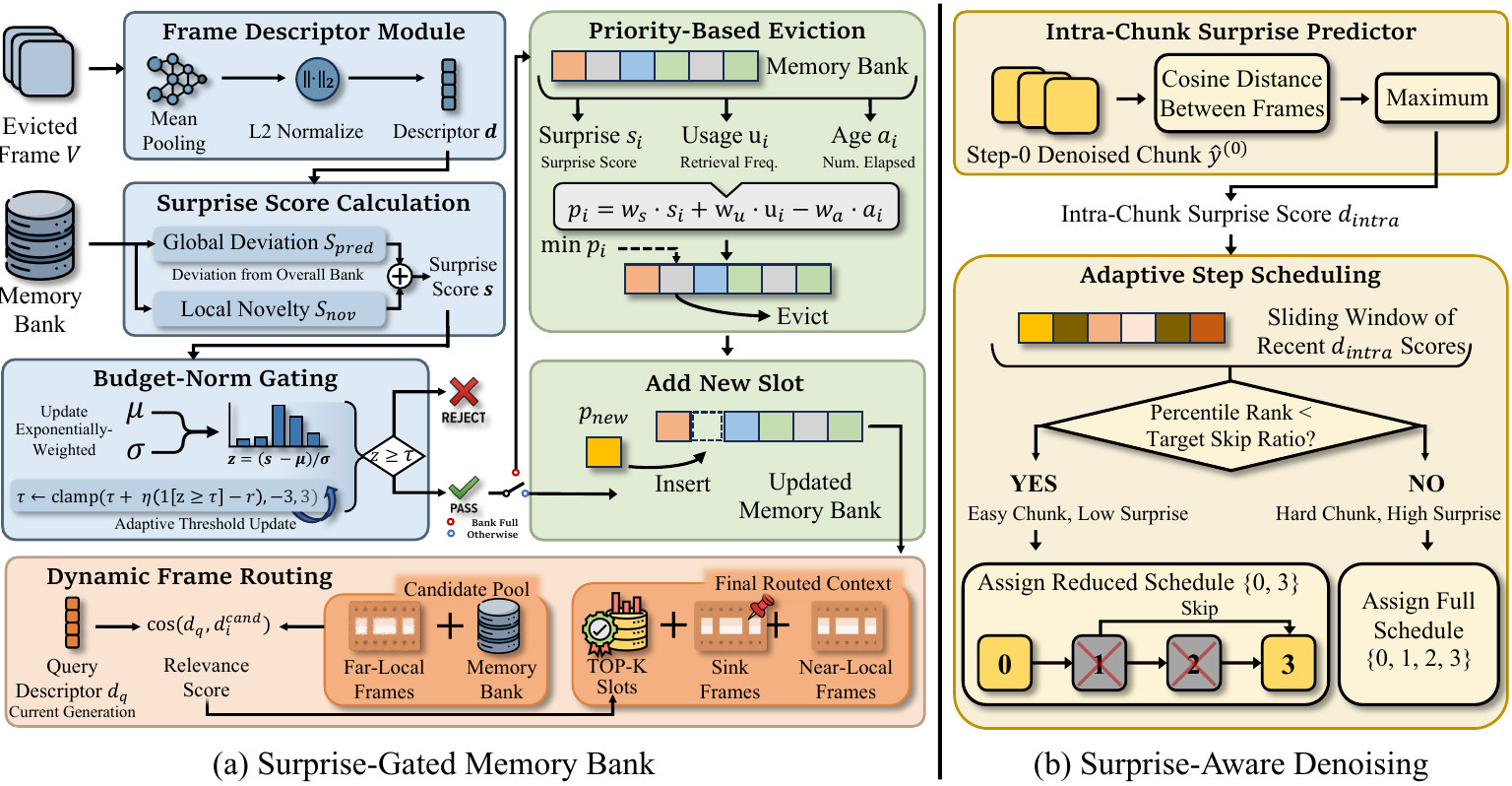}
    \caption{The framework of Surprise Forcing. Our method consists of two surprise-driven components. The first is the Surprise-Gated Memory Bank, which regulates the writing and eviction of memory entries based on surprise-related signals. The second is Surprise-Aware Denoising, which employs an Intra-Chunk Surprise Predictor to estimate whether the denoising process for the current chunk can be accelerated.}
    \label{fig:framework}
\end{figure}

Figure \ref{fig:framework} summarizes the two controllers. The memory path observes frames when they leave the local cache, decides whether they contain information not already represented by the bank, and manages a bounded set of accepted entries. At attention time, it combines bank entries and still-valid far-local frames, selecting only those that match the current query. The denoising path observes the partially denoised current chunk, estimates its relative difficulty, and either keeps or shortens the remaining schedule. Both paths are causal and require no parameter update.

\subsection{Surprise-Gated Memory Bank}
\label{sec:memory_bank}

A chronological cache treats all history identically until eviction. The external bank instead asks whether a departing frame expands the representation of the past. Static duplicates should usually be rejected; a new pose, viewpoint, object configuration, or scene transition should be more likely to survive.

\noindent\textbf{Frame descriptor.}
For an evicted frame, we summarize its spatial value tokens and normalize the result:

\begin{equation}
\bar{\mathbf{v}} = \frac{1}{L}\sum_{l=1}^{L} \mathbf{V}_{l} \in \mathbb{R}^{D}, \quad
\mathbf{d} = \frac{\bar{\mathbf{v}}}{\|\bar{\mathbf{v}}\|_2}.
\end{equation}

Value tokens are preferable to key tokens for content comparison in this setting. Keys are optimized for query-dependent discrimination and interact with positional encoding, which can make similarity sensitive to attention routing rather than visual content. Values carry the information propagated through the attention output and empirically provide a more stable content descriptor.

\noindent\textbf{Dual-component surprise.}
A single bank similarity is insufficient. Mean similarity can hide a near-duplicate of one stored frame, whereas nearest-neighbor similarity can overlook whether the candidate is already covered by the bank distribution as a whole. With bank descriptors $\{\mathbf{d}_i\}_{i=1}^{M}$, we therefore use a global deviation term $s_{\mathrm{pred}}=\frac{1}{2}(1-\frac{1}{M}\sum_i\cos(\mathbf{d},\mathbf{d}_i))$ and a local novelty term $s_{\mathrm{nov}}=\frac{1}{2}(1-\max_i\cos(\mathbf{d},\mathbf{d}_i))$. Their mixture is

\begin{equation}
s = \alpha \cdot s_{\text{pred}} + (1 - \alpha) \cdot s_{\text{nov}}
\end{equation}

\noindent\textbf{Budget-Norm Gating.}
Raw surprise is not calibrated across videos or even across phases of the same video. A static segment may produce scores in a narrow low range, while an action segment shifts the entire distribution upward. A fixed raw or fixed normalized threshold therefore produces unstable bank utilization. Budget-Norm Gating separates \emph{relative unusualness} from the desired write frequency.

For each bank, we maintain exponentially weighted online statistics. Using time-indexed notation,
\begin{equation}
\mu_{t+1}=m\mu_t+(1-m)s_t,
\label{eq:ema_mean}
\end{equation}
\begin{equation}
\sigma_{t+1}^{2}=m\sigma_t^{2}+(1-m)(s_t-\mu_{t+1})^{2},
\label{eq:ema_var}
\end{equation}
and normalize the current score as
\begin{equation}
z_t=\frac{s_t-\mu_t}{\sigma_t}, \qquad \sigma_t=\sqrt{\sigma_t^2}.
\label{eq:zscore}
\end{equation}
The statistics are updated for every evaluated frame, not only accepted frames, so they track the uncensored local score distribution.

Admission is $\mathrm{pass}_t=\mathbf{1}[z_t\geq\tau_t]$. The threshold follows the feedback rule
\begin{equation}
\tau_{t+1}=\mathrm{clamp}\!\left(\tau_t+\eta(\mathrm{pass}_t-r),-3,3\right),
\label{eq:budget_controller}
\end{equation}
where $r$ is the target admission ratio. Ignoring the clamp, the expected threshold drift is $\eta(\Pr[\mathrm{pass}_t=1]-r)$; the stationary condition is therefore an acceptance probability of $r$. The controller raises the bar after excessive writes and lowers it when admission is too sparse, making the budget responsive to content rather than score scale.

During cold start, the gate is bypassed for the first few evaluations so that the bank and running statistics can be populated. The initial controller state is $\tau_0=0.002$ in our experiments; while the bank is under-filled, the active threshold is capped by this value to prevent premature starvation. Clamping to $[-3,3]$ avoids a saturated state that would accept or reject almost everything. The target $r$ describes the surprise gate itself: the final commit rate can be lower because near-duplicate filtering and neighboring-chunk throttling may reject a candidate after it passes the controller.

\begin{figure}[t]
    \centering
    \includegraphics[width=\linewidth]{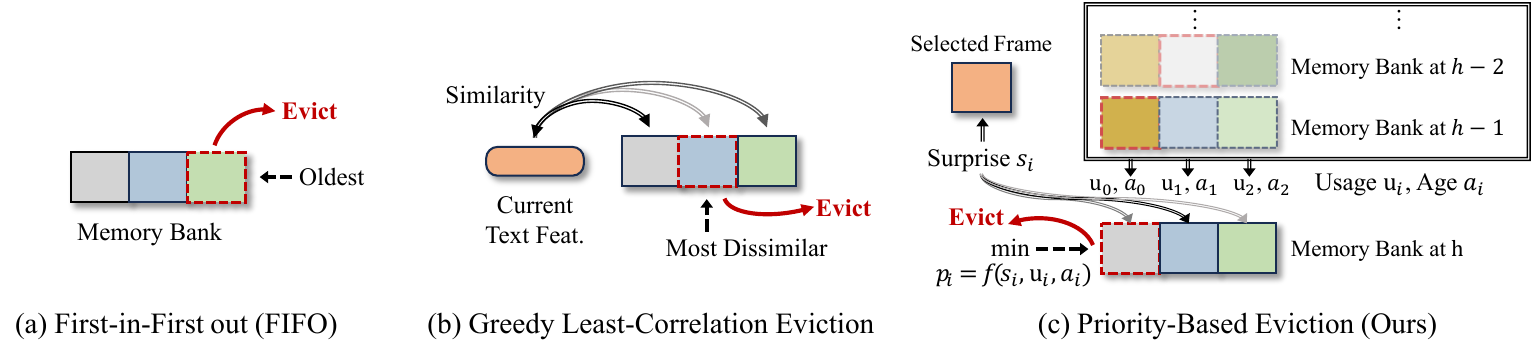}
    \caption{Comparison of historical context eviction strategies. Unlike previous methods that rely solely on temporal order or current relevance as the eviction criterion, our approach provides a more principled mechanism for memory eviction.
}
    \label{fig:evict}
\end{figure}

\subsubsection{Priority-Based Eviction.}
When the bank reaches capacity, replacement should not be determined by age alone. An old establishing view may still be repeatedly retrieved, and a currently low-relevance frame may be the only remaining representative of an earlier state. Conversely, a recent entry may be redundant with the active sliding window. We score each slot using its write-time surprise, accumulated retrieval usage, and age:

\begin{equation}
p_i = w_s \cdot s_i + w_u \cdot u_i - w_a \cdot a_i
\end{equation}

All three signals are normalized to $[0,1]$. The pending frame receives the same score, and it replaces the minimum-priority entry only when its priority is larger. The usage term protects entries that remain operationally useful, while the age penalty prevents the bank from becoming permanently occupied by early frames.

\subsubsection{Dynamic Frame Routing.}
Selective writing does not imply that every stored frame should participate in every attention operation. Attending to the complete bank increases cost and can dilute the near-local signal. For each chunk, we therefore form a candidate pool from valid bank entries and far-local frames that remain in the rolling cache but lie outside the near-local neighborhood. A descriptor $\mathbf{d}_q$ for the current generation query scores candidate $i$ by $\cos(\mathbf{d}_q,\mathbf{d}^{\mathrm{cand}}_i)$. The top-$k$ candidates are routed together with sink and near-local frames. This preserves broad temporal coverage while keeping the global attention budget fixed.

\subsection{Surprise-Aware Denoising}
\label{sec:early_exit}

\begin{figure}[t]
    \centering
    \includegraphics[width=\linewidth]{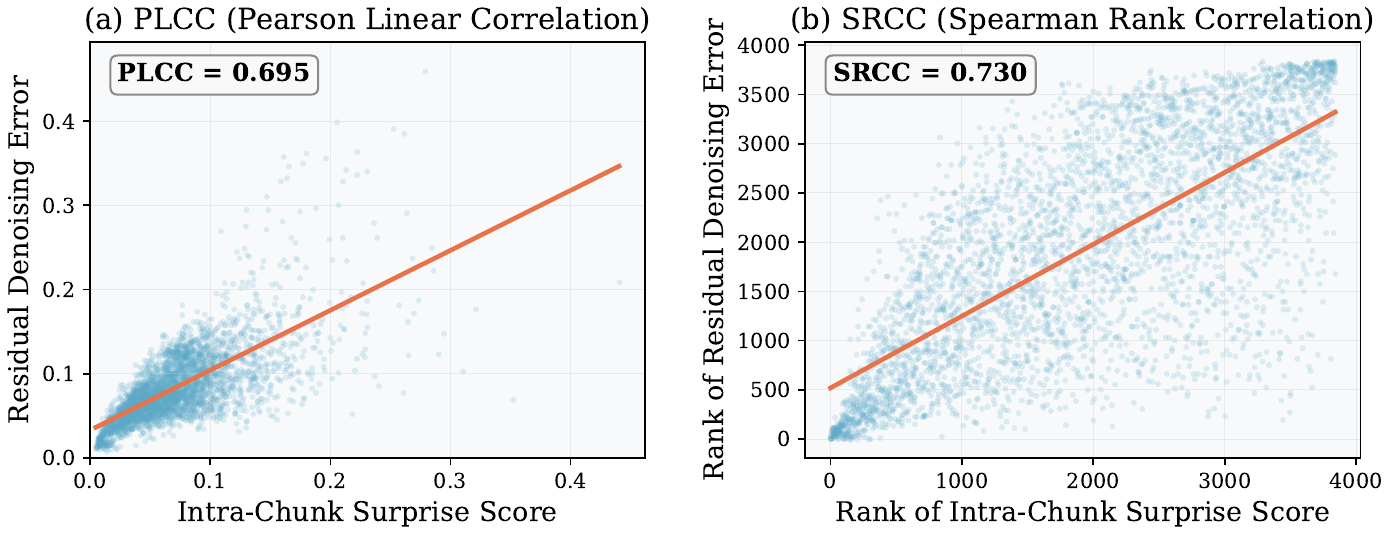}
    \caption{The statistical correlation between the intra-chunk surprise score and the MSE between the first and the final denoising steps of each chunk. As illustrated in (a) and (b), the two variables exhibit a clear positive relationship, which is visually evident from the strong upward trend in both plots.}
    \label{fig:correlation}
\end{figure}

A four-step distilled sampler executes four transformer passes for every chunk, even when the output is already close to its final state after the first pass. We seek a signal that is available early, does not require an auxiliary predictor, and remains meaningful as the content changes.

\noindent\textbf{Intra-chunk surprise predictor.}
After the first denoising pass, the partially denoised latent $\hat{y}^{(0)}\!\in\!\mathbb{R}^{B\times T\times C\times H\times W}$ reveals the temporal variation within the current chunk. We compute adjacent-frame cosine distances and retain the largest transition:

\begin{gather}
c_t = 1 - \cos\!\left(\text{vec}(\hat{y}^{(0)}_{t}),\ \text{vec}(\hat{y}^{(0)}_{t+1})\right), \quad t = 1, \ldots, T{-}1 \\
d_{\text{intra}} = \max_{t}\ c_t
\end{gather}

The maximum is used because a single abrupt transition can determine the amount of remaining refinement even when the other frame pairs are nearly static. Cosine distance removes sensitivity to the changing global latent magnitude. On 50 MovieGen prompts with one-minute generations, this score has PLCC $0.695$ and SRCC $0.730$ with the residual error between the first and final denoising outputs, as visualized in \cref{fig:correlation}. The signal is self-referential: it depends only on the current chunk's first-step output and therefore avoids dependence on a separate learned calibration model.

\noindent\textbf{Adaptive step scheduling.}
Difficulty is judged relative to recent chunks rather than by a fixed global threshold. We rank $d_{\mathrm{intra}}$ within a sliding window and use

\begin{equation}
\text{steps} = \begin{cases} 
\{0, 3\} & \text{if rank} < r_{skip} \quad (\text{skip intermediate steps}) \\ 
\{0, 1, 2, 3\} & \text{otherwise} \quad (\text{full schedule}) 
\end{cases}
\end{equation}

where $r_{skip}$ controls the fraction of comparatively easy chunks eligible for the reduced schedule. In the implementation, each chunk contains $T=3$ frames, the ranking window contains 20 recent chunks, and the first five chunks always use all four steps. Thus, the scheduler first builds a local reference distribution and then adapts to the current video's motion regime.

\section{Experiments}

\subsection{Experimental Setup}
\noindent\textbf{Base model and generation protocol.}
We implement Surprise Forcing on LongLive~\cite{yang2025longlive}, a Wan2.1-T2V-1.3B~\cite{wan2025wan} streaming generator trained with a Self Forcing DMD pipeline. The model produces short causal chunks conditioned on the rolling cache. Unless stated otherwise, short-video evaluation uses 5-second outputs and long-video evaluation uses 60-second outputs at $832\times480$ resolution.

\noindent\textbf{Memory configuration.}
The bank capacity is $C=6$ for long videos and $C=3$ for short videos. We use $\alpha=0.7$, EMA momentum $m=0.95$, controller step size $\eta=0.1$, target admission ratio $r=0.3$, and $\tau_0=0.002$. Priority weights are $\{w_s,w_u,w_a\}=\{1.8,1.0,0.4\}$, and routing selects $k=3$ candidates. The first three chunks are admitted during memory warmup; their stored surprise values are subsequently reset to the current EMA mean so that the initial population does not receive an artificial priority advantage.

\noindent\textbf{Denoising configuration.}
The full schedule is $\{0,1,2,3\}$ and the reduced schedule is $\{0,3\}$. Our default is $r_{skip}=0.4$. The percentile window and warmup follow the settings described in \cref{sec:early_exit}.

\noindent\textbf{Metrics.}
We use the official prompts and metrics from VBench~\cite{huang2024vbench} and VBench-Long~\cite{huang2025vbench++}. We report VBench Total score, VBench-Long Quality score, representative semantic and visual dimensions, and throughput in frames per second. Long-range consistency is evaluated with VBench-2.0~\cite{zheng2025vbench2}. Component studies that require a larger set of long videos use 128 randomly sampled MovieGen~\cite{polyak2024movie} prompts.

\subsection{Comparison with Streaming Alternatives}
We compare with CausVid~\cite{yin2025slow}, Self Forcing~\cite{huang2025self}, Rolling Forcing~\cite{liu2025rolling}, and LongLive~\cite{yang2025longlive}. These methods share the broad objective of efficient causal generation but differ in distillation, rollout, scheduling, and context handling.

\begin{table}[H]
\centering
\caption{Quantitative comparisons on VBench(5s)~\cite{huang2024vbench} short videos and VBench-Long(60s)~\cite{huang2025vbench++} long videos. Under comparable inference speed, our method achieves state-of-the-art performance across all metrics, and the speedup becomes more pronounced when generating longer videos.
}
\tablebodyfont
\setlength{\tabcolsep}{1.6mm}
\label{tab:main_result}
\begin{tabular}{lcccccccc}
\toprule
\multirow{2}{*}{\tablehead{Model}} & \multicolumn{4}{c}{\tablehead{VBench}} & \multicolumn{4}{c}{\tablehead{VBench-Long}} \\
\cmidrule(lr){2-5} \cmidrule(lr){6-9}
 & \tablehead{Total}$\uparrow$ & \makecell{\tablehead{Sub.}\\\tablehead{Consistency} $\uparrow$} & \makecell{\tablehead{Mul.}\\\tablehead{Objects} $\uparrow$} & \tablehead{FPS} $\uparrow$ & \tablehead{Quality}$\uparrow$ & \makecell{\tablehead{Sub.}\\\tablehead{Consistency} $\uparrow$} & \tablehead{Aesthetic} $\uparrow$ & \tablehead{FPS} $\uparrow$ \\
\midrule
CausVid \cite{yin2025slow} & 80.47 & 94.66 & 79.99 & 15.84 & 75.97 & 96.41 & 48.42 & 15.84 \\
Self Forcing \cite{huang2025self} & 82.12 & 96.23 & 80.22 & 15.84 & 80.53 & 97.19 & 55.98 & 15.84 \\
Rolling Forcing \cite{liu2025rolling} & 80.49 & 96.21 & 84.53 & 15.52 & 80.23 & 98.3 & 60.97 & 15.52 \\
LongLive \cite{yang2025longlive} & 80.63 & 96.19 & 84.89 & \textbf{18.01} & 80.32 & 98.42 & 61.71 & \textbf{18.01} \\
Surprise Forcing & \textbf{83.40} & \textbf{96.28} & \textbf{87.50} & 16.45 & \textbf{81.88} & \textbf{98.51} & \textbf{63.98} & 17.18 \\
\bottomrule
\end{tabular}
\end{table}

\noindent\textbf{Quantitative results.}
On 5-second VBench generations, Surprise Forcing improves the Total score from the strongest baseline value of $82.12$ to $83.40$ and raises Multiple Objects to $87.50$. The gain is obtained without changing the trained network. On 60-second VBench-Long generations, the method reaches the best reported Quality, Subject Consistency, and Aesthetic scores in \cref{tab:main_result}, while running at $17.18$ FPS. The result supports the central hypothesis that selective global history can improve long-horizon quality without forfeiting streaming operation.

\begin{figure}[t]
    \centering
    \includegraphics[width=\linewidth]{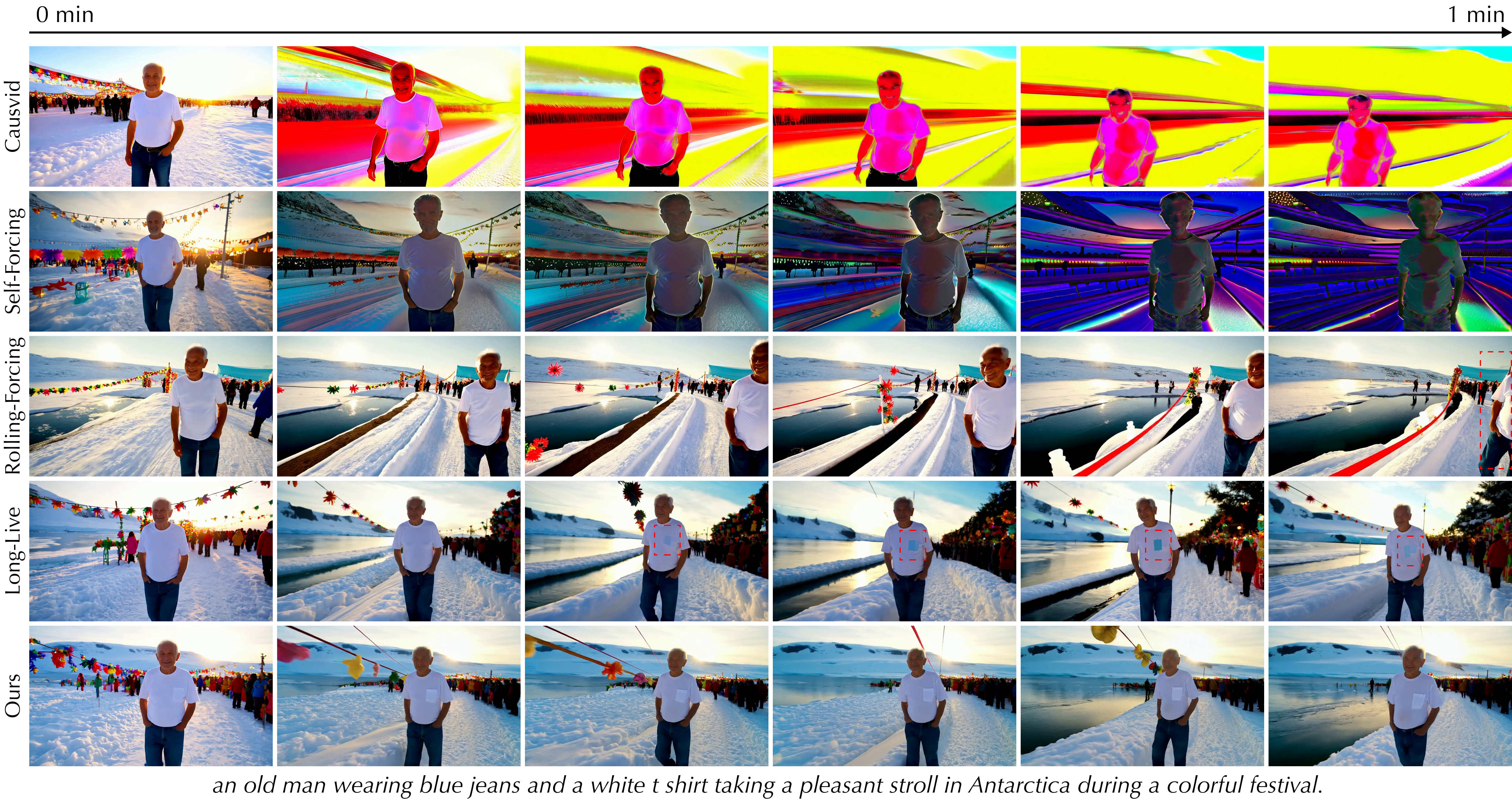}
    \caption{Qualitative comparison with other methods. Our approach maintains strong subject consistency and smooth motion continuity during long video generation, while introducing minimal degradation in visual quality.}
    \label{fig:comparison}
\end{figure}

{\clubpenalty=10000
\noindent\textbf{Qualitative comparison.} The one-minute example in \cref{fig:comparison} exposes different modes of long-horizon degradation. CausVid accumulates color and quality errors; Self Forcing deteriorates after moving beyond its effective training horizon; Rolling Forcing improves visual stability but allows the subject to drift spatially; and LongLive retains overall quality while subject details change over time. Surprise Forcing better preserves subject appearance and motion continuity because historically distinctive states remain available after they leave the local cache.\par}

\subsection{Targeted Long-Range Consistency}
Aggregate benchmark scores can obscure the particular failures caused by forgetting. We therefore report four VBench-2.0 dimensions chosen to match the intended role of the memory bank: Human Identity, Human Clothes, Instance Preservation, and Multi-View Consistency.

\begin{table}[H]
\setlength\tabcolsep{8pt}
\renewcommand{\arraystretch}{1.2}   
  \centering
\setlength{\belowcaptionskip}{1pt}
  \setlength{\abovecaptionskip}{5pt}
  \caption{Long-range consistency results on selected VBench-2.0 metrics.}
  \tablebodyfont
  \begin{tabular}{@{}lccccc@{}}
    \toprule
    \tablehead{Metric} & \tablehead{CausVid} & \tablehead{Self Forcing} & \tablehead{Rolling Forcing} & \tablehead{LongLive} & \tablehead{Surprise Forcing} \\
    \midrule                          
    \textbf{Human Identity}         & 63.96 & 65.81 & 61.94 & 78.23 & \textbf{82.63} \\
    \textbf{Human Clothes}          & 93.33 & 93.84 & 93.24 & 95.27 & \textbf{96.67} \\
    \textbf{Instance Preservation}  & 71.93 & 77.19 & 77.19 & 79.53 & \textbf{83.63} \\
    \textbf{Multi-View Consistency} & 23.80 & 24.10 & 18.87 & 24.53 & \textbf{29.76} \\
    \bottomrule
  \end{tabular}
  \label{tab:vbench2}
\end{table}

The method is strongest on all four dimensions. The largest absolute improvement over LongLive occurs for Multi-View Consistency, where the score increases from $24.53$ to $29.76$. Identity and instance preservation also improve by more than four points. These measurements complement the main benchmark by directly testing whether information from earlier views and appearances remains recoverable.

\subsection{Core Controller Ablations}
The compact ablations in \cref{tab:gate_ablation,tab:eviction_ablation,tab:sad_ablation} isolate the three online controllers on the official 30-second VBench-Long prompt set.

\begin{table}[H]
\centering
\setlength{\tabcolsep}{8pt}
\caption{Ablation study of fixed gating and budget-norm gating.}
\tablebodyfont
\label{tab:gate_ablation}
\begin{tabular}{lccc}
\toprule
\tablehead{Method} & \begin{tabular}[c]{@{}c@{}}\tablehead{Total}\\\tablehead{Score} $\uparrow$\end{tabular} & \begin{tabular}[c]{@{}c@{}}\tablehead{Quality}\\\tablehead{Score} $\uparrow$\end{tabular} & \begin{tabular}[c]{@{}c@{}}\tablehead{Semantic}\\\tablehead{Score} $\uparrow$\end{tabular} \\ \midrule
$z_{thres}=0.1$     & 81.24 & 81.37 & 80.7 \\ 
$z_{thres}=2.0$     & 81.66 & 81.83 & 80.98\\
\textbf{Budget-Norm}  & \textbf{83.20} & \textbf{83.70} & \textbf{81.21} \\
\bottomrule
\end{tabular}
\end{table}

\begin{table}[H]
\centering
\setlength{\tabcolsep}{8pt}
\caption{Ablation study of FIFO and priority-based eviction.}
\tablebodyfont
\label{tab:eviction_ablation}
\begin{tabular}{lccc}
\toprule
\tablehead{Method} & \begin{tabular}[c]{@{}c@{}}\tablehead{Total}\\\tablehead{Score} $\uparrow$\end{tabular} & \begin{tabular}[c]{@{}c@{}}\tablehead{Quality}\\\tablehead{Score} $\uparrow$\end{tabular} & \begin{tabular}[c]{@{}c@{}}\tablehead{Semantic}\\\tablehead{Score} $\uparrow$\end{tabular} \\ \midrule
FIFO   & 82.92 & 83.44 & 80.84 \\
\textbf{Priority}     & \textbf{83.20} & \textbf{83.70} & \textbf{81.21} \\ \bottomrule
\end{tabular}
\end{table}

\begin{table}[t]
\centering
\setlength{\tabcolsep}{7pt}
\caption{Ablation study of different skip ratios.}
\tablebodyfont
\label{tab:sad_ablation}
\begin{tabular}{lcccc}
\toprule
\tablehead{Skip Ratio}
& \begin{tabular}[c]{@{}c@{}}\tablehead{Total}\\\tablehead{Score} $\uparrow$\end{tabular}
& \begin{tabular}[c]{@{}c@{}}\tablehead{Quality}\\\tablehead{Score} $\uparrow$\end{tabular}
& \begin{tabular}[c]{@{}c@{}}\tablehead{Semantic}\\\tablehead{Score} $\uparrow$\end{tabular}
& \tablehead{Acc. Ratio} \\
\midrule
$r_{skip}=0.2$   & 83.36 & 83.85 & 81.40 & 8.7\% \\
$r_{skip}=0.4$   & 83.20 & 83.70 & 81.21 & 15.8\% \\
$r_{skip}=0.6$   & 82.15 & 82.46 & 80.93 & 21.6\% \\
$r_{skip}=0$     & 84.03 & 84.55 & 81.95 & 0\% \\ 
\bottomrule
\end{tabular}
\end{table}

\noindent\textbf{Admission controller.}
Replacing feedback control with a fixed normalized threshold hurts all reported scores. A permissive threshold writes too many low-value frames and makes routing less selective; a conservative threshold leaves the bank under-utilized. Budget-Norm Gating avoids both regimes by adapting threshold strictness to the observed stream.

\noindent\textbf{Replacement policy.}
FIFO discards frames solely because they are old. Priority-based eviction instead preserves entries that were surprising or remain frequently retrieved, producing higher Total, Quality, and Semantic scores.

\noindent\textbf{Step allocation.}
Increasing $r_{skip}$ yields a monotonic acceleration--quality trade-off. The default $r_{skip}=0.4$ reduces denoising work by $15.8\%$ while remaining close to the full-schedule scores. At $r_{skip}=0.6$, acceleration reaches $21.6\%$, but the quality reduction becomes more visible.

\subsection{Surprise Score Decomposition}
The memory score combines distribution-level deviation and nearest-neighbor novelty. We test each term alone, the default prediction-dominant mixture, and a novelty-dominant swapped mixture on 128 one-minute MovieGen generations.

\begin{table}[H]
\centering
\setlength{\tabcolsep}{5pt}
\caption{Quantitative comparison between different surprise score settings. Here, $S_{\text{swap}}$ denotes a variant where the mixing weights of $S_{\text{pred}}$ and $S_{\text{nov}}$ in the main text are swapped, in order to investigate the case where the Surprise Score places greater emphasis on $S_{\text{nov}}$. All generated videos are 60 seconds in length.}
\tablebodyfont
\label{tab:surprise_score}
\begin{tabular}{lcccccc}
\toprule
\tablehead{Method} &
\begin{tabular}[c]{@{}c@{}}\tablehead{Subject}\\\tablehead{Consistency}\end{tabular} &
\begin{tabular}[c]{@{}c@{}}\tablehead{Background}\\\tablehead{Consistency}\end{tabular} &
\begin{tabular}[c]{@{}c@{}}\tablehead{Motion}\\\tablehead{Smoothness}\end{tabular} &
\begin{tabular}[c]{@{}c@{}}\tablehead{Dynamic}\\\tablehead{Degree}\end{tabular} &
\begin{tabular}[c]{@{}c@{}}\tablehead{Aesthetic}\\\tablehead{Quality}\end{tabular} &
\begin{tabular}[c]{@{}c@{}}\tablehead{Imaging}\\\tablehead{Quality}\end{tabular} \\
\midrule
    
$S_{pred}$ & 97.12 & 95.94 & 98.39 & 55.75 & 61.67 & 68.30 \\
$S_{nov}$ & 97.07 & 95.96 & 98.35 & 55.32 & 61.46 & 68.11 \\
$S_{swap}$ & 97.13 & 96.02 & 98.37 & 56.28 & \textbf{61.96} & 68.47 \\
\textbf{Ours}        & \textbf{97.16} & \textbf{96.25} & \textbf{98.44} & \textbf{56.56} & 61.63 & \textbf{68.49} \\
\bottomrule
\end{tabular}
\end{table}

Using only $S_{\mathrm{pred}}$ is slightly stronger than using only $S_{\mathrm{nov}}$, indicating that deviation from the bank as a whole is the more reliable base signal. Nevertheless, both mixtures outperform the single terms across several dimensions. The novelty term is therefore useful as a correction for sparse events that mean aggregation can dilute. We retain $0.7S_{\mathrm{pred}}+0.3S_{\mathrm{nov}}$ because it gives the best overall balance rather than optimizing a single metric.

\subsection{Memory Capacity, Routing, and Admission Budget}
\noindent\textbf{Bank size.}

\begin{table}[H]
\setlength\tabcolsep{2pt}
\renewcommand{\arraystretch}{1.2}        
  \centering
  \setlength{\belowcaptionskip}{1pt}
  \setlength{\abovecaptionskip}{5pt}
  \caption{Ablation analysis for bank size.}
  \tablebodyfont
  \begin{tabular}{@{}lcccccccc@{}}
    \toprule
    \tablehead{Bank Size} &
      \makecell{\tablehead{Subject}\\\tablehead{Cons.}} &
      \makecell{\tablehead{Backgr.}\\\tablehead{Cons.}} &
      \makecell{\tablehead{Motion}\\\tablehead{Smooth.}} &
      \makecell{\tablehead{Dynamic}\\\tablehead{Degree}} &
      \makecell{\tablehead{Aesthetic}\\\tablehead{Quality}} &
      \makecell{\tablehead{Imaging}\\\tablehead{Quality}} &
      \tablehead{FPS} &
      \makecell{\tablehead{GPU}\\\tablehead{Mem.}} \\
    \midrule
    $c=3$  & 97.04 & 96.13 & 98.42 & 56.50 & 61.42 & 68.33 & 17.33 & 22.67 \\
    $c=6$  & 97.16 & 96.25 & 98.44 & 56.56 & 61.63 & 68.49 & 17.18 & 23.48 \\
    $c=9$  & 97.23 & 96.22 & 98.46 & 56.98 & 61.60 & 68.46 & 16.78 & 24.28 \\
    $c=12$ & 97.28 & 96.29 & 98.47 & 54.79 & 61.75 & 68.63 & 16.70 & 25.08 \\
    \bottomrule
  \end{tabular}
  \label{tab:bank}
\end{table}

Increasing capacity from $3$ to $12$ produces only small and inconsistent quality changes while lowering FPS and increasing GPU memory. Capacity $C=6$ lies near the favorable part of this curve: it retains enough temporal diversity for long videos without turning the external bank into a large attention burden.

\noindent\textbf{Routing strategy.}

\begin{table}[H]
\setlength\tabcolsep{2pt}
\renewcommand{\arraystretch}{1.2}        
  \centering
  \setlength{\belowcaptionskip}{1pt}
  \setlength{\abovecaptionskip}{5pt}
  \caption{Ablation analysis for dynamic routing method.}
  \tablebodyfont
  \begin{tabular}{@{}lcccccccc@{}}
    \toprule
    \tablehead{Routing} &
      \makecell{\tablehead{Subject}\\\tablehead{Cons.}} &
      \makecell{\tablehead{Backgr.}\\\tablehead{Cons.}} &
      \makecell{\tablehead{Motion}\\\tablehead{Smooth.}} &
      \makecell{\tablehead{Dynamic}\\\tablehead{Degree}} &
      \makecell{\tablehead{Aesthetic}\\\tablehead{Quality}} &
      \makecell{\tablehead{Imaging}\\\tablehead{Quality}} &
      \tablehead{FPS} &
      \makecell{\tablehead{GPU}\\\tablehead{Mem.}} \\
    \midrule
    $k=3$    & 97.16 & 96.25 & 98.44 & 56.56 & 61.63 & 68.49 & 17.18 & 23.48 \\
    $k=6$    & 97.24 & 96.29 & 98.57 & 54.42 & 61.48 & 68.01 & 16.14 & 24.26 \\
    recent   & 96.25 & 95.34 & 97.32 & 54.84 & 61.27 & 68.25 & 17.18 & 23.48 \\
    random   & 96.23 & 95.16 & 97.49 & 54.73 & 61.44 & 68.28 & 17.18 & 23.48 \\
    \bottomrule
  \end{tabular}
  \label{tab:routing}
\end{table}

Selecting six frames slightly improves a few consistency measures but reduces dynamic degree, aesthetic quality, imaging quality, and throughput. Random or recency-only selection is substantially worse than similarity-based top-$3$ routing. The result indicates that the value of the bank comes not only from storing history but also from limiting retrieval to context that is relevant to the current query.

\noindent\textbf{Target admission ratio.}

\begin{table}[H]
\setlength\tabcolsep{5pt}
\renewcommand{\arraystretch}{1.2}        
  \centering
  \setlength{\belowcaptionskip}{1pt}
  \setlength{\abovecaptionskip}{5pt}
  \setlength{\heavyrulewidth}{0.05em}
  \setlength{\lightrulewidth}{0.03em}
  \caption{Ablation analysis for target write ratio.}
  \tablebodyfont
  \begin{tabular}{@{}lcccccc@{}}
    \toprule
    \makecell{\tablehead{Write}\\\tablehead{Ratio}} &
    \makecell{\tablehead{Subject}\\\tablehead{Cons.}} &
    \makecell{\tablehead{Backgr.}\\\tablehead{Cons.}} &
    \makecell{\tablehead{Motion}\\\tablehead{Smooth.}} &
    \makecell{\tablehead{Dynamic}\\\tablehead{Degree}} &
    \makecell{\tablehead{Aesthetic}\\\tablehead{Quality}} &
    \makecell{\tablehead{Imaging}\\\tablehead{Quality}} \\
    \midrule
    $r=0.1$ & 97.09 & 96.09 & 98.38 & 55.16 & 61.58 & 68.43 \\
    $r=0.3$ & \textbf{97.16} & \textbf{96.25} & \textbf{98.44} & \textbf{56.56} & \textbf{61.63} & \textbf{68.49} \\
    $r=0.5$ & 97.12 & 96.14 & 98.39 & 54.84 & 61.51 & 68.42 \\
    $r=0.7$ & 97.09 & 96.09 & 98.37 & 54.74 & 61.59 & 68.39 \\
    \bottomrule
  \end{tabular}
  \label{tab:write}
\end{table}

The best overall setting is $r=0.3$. At $r=0.1$, the controller admits too little history; at $r=0.5$ or $0.7$, frequent turnover removes useful entries before they can support later routing. This experiment clarifies that the target is a stability--freshness operating point, not simply a request to maximize writes.

\subsection{Alternative Eviction and Denoising Predictors}
\noindent\textbf{Greedy least-correlation eviction.}

\begin{table}[H]
\centering
\setlength{\tabcolsep}{5pt}
\caption{Quantitative comparison between our eviction method and Greedy Least-Correlation Eviction~\cite{ji2025memflow}. All generated videos are 60 seconds in length.}
\tablebodyfont
\label{tab:text_cos_eviction}
\begin{tabular}{lcccccc}
\toprule
\tablehead{Method} &
\begin{tabular}[c]{@{}c@{}}\tablehead{Subject}\\\tablehead{Consistency}\end{tabular} &
\begin{tabular}[c]{@{}c@{}}\tablehead{Background}\\\tablehead{Consistency}\end{tabular} &
\begin{tabular}[c]{@{}c@{}}\tablehead{Motion}\\\tablehead{Smoothness}\end{tabular} &
\begin{tabular}[c]{@{}c@{}}\tablehead{Dynamic}\\\tablehead{Degree}\end{tabular} &
\begin{tabular}[c]{@{}c@{}}\tablehead{Aesthetic}\\\tablehead{Quality}\end{tabular} &
\begin{tabular}[c]{@{}c@{}}\tablehead{Imaging}\\\tablehead{Quality}\end{tabular} \\
\midrule
    
Greedy Least-Correlation Eviction & 97.14 & 96.07 & 98.36 & 55.00 & \textbf{61.64} & 68.37 \\
\textbf{Priority-Based Eviction}        & \textbf{97.16} & \textbf{96.25} & \textbf{98.44} & \textbf{56.56} & 61.63 & \textbf{68.49} \\
\bottomrule
\end{tabular}
\end{table}

Evicting according to the current text-retrieval demand is myopic: an entry that is not useful now can become important after a later viewpoint or narrative return. The priority policy is stronger on consistency, motion, and imaging quality because it incorporates historical surprise, actual usage, and age rather than a single instantaneous query.

\noindent\textbf{Chunk-difficulty proxy.}

\begin{table}[H]
\centering
\setlength{\tabcolsep}{6pt}
\caption{Correlation comparison of different intra-chunk predictors.}
\tablebodyfont
\label{tab:predictor_corr}
\begin{tabular}{lcc}
\toprule
\tablehead{Predictor} & \tablehead{PLCC} & \tablehead{SRCC} \\
\midrule
First-Last Surprise Score & 0.6641 & 0.7123 \\
\textbf{Inter-frame Max Surprise Score}   & \textbf{0.6953} & \textbf{0.7298} \\
\bottomrule
\end{tabular}
\end{table}

A first--last-frame score measures only net endpoint change and can miss an abrupt transition in the middle of a three-frame chunk. The maximum adjacent-frame score better tracks residual denoising error, improving both PLCC and SRCC, and is therefore used in all experiments.

\subsection{Qualitative Component Analysis}

\begin{figure}[t]
    \centering
    \includegraphics[width=\linewidth]{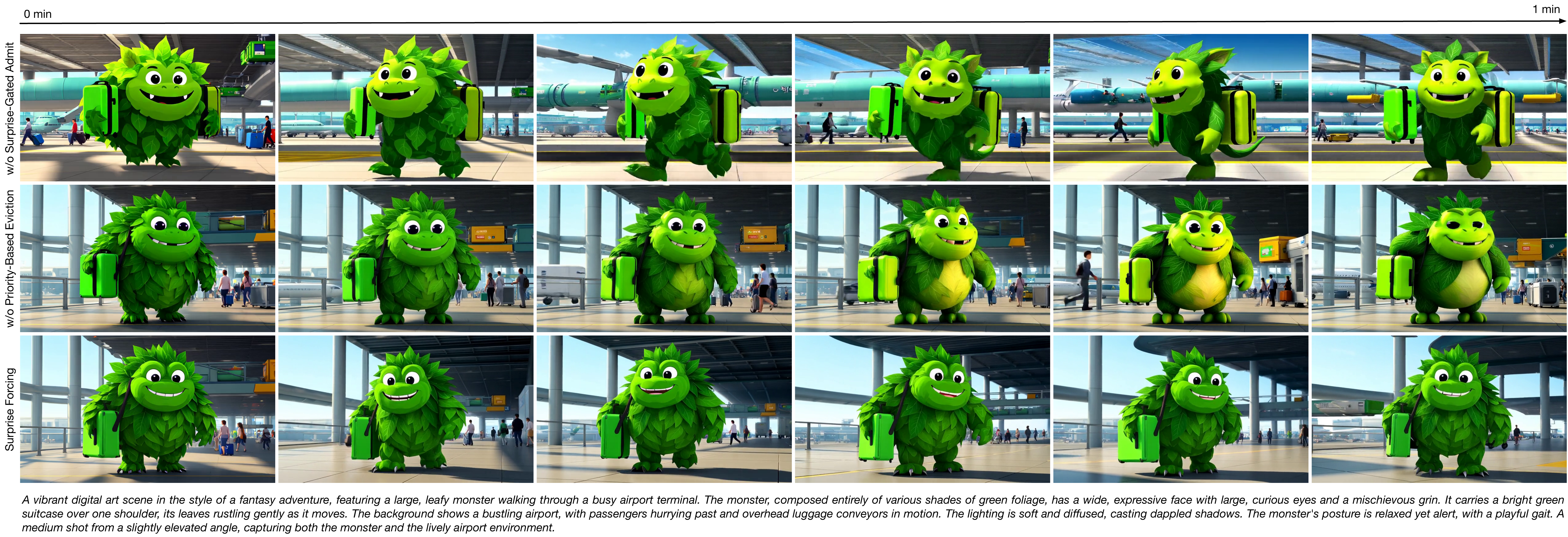}
    \caption{Qualitative analysis of the proposed Surprise-Gated Memory Bank.}
    \label{fig:ablation}
\end{figure}

The ablation in \cref{fig:ablation} visualizes why both memory admission and replacement matter. Writing evicted frames chronologically without surprise gating allows redundant or weak entries to occupy the bank, after which subject identity and frame quality progressively degrade. Replacing priority-based eviction with FIFO causes attributes such as color and foliage structure to drift because informative older states are removed prematurely. The full method maintains a cleaner historical basis for later chunks.

\section{Discussion and Limitations}
Surprise Forcing demonstrates that useful control signals can be extracted from the same inference trajectory that they regulate. The memory controller uses relative novelty rather than a model-specific absolute score, and the denoising controller uses local percentile rank rather than a globally calibrated threshold. This design helps the framework adapt across static and dynamic segments while remaining training-free.

Several limitations remain. First, the descriptor is a compact mean-pooled summary and may not distinguish spatial arrangements that share similar global content. Second, the priority weights and target ratios are fixed across a rollout; a higher-level controller could adapt these budgets to task or prompt structure. Third, the current evaluation focuses on clips up to one minute and does not establish behavior over substantially longer or interactive trajectories. Finally, step skipping is binary between two schedules. A richer controller could assign a broader family of solver paths while explicitly estimating uncertainty.

\section{Conclusion}
We present Surprise Forcing as a unified allocation mechanism for streaming long-video diffusion. The Surprise-Gated Memory Bank preserves a small, useful subset of evicted history through normalized surprise, feedback-controlled admission, priority replacement, and dynamic routing. Surprise-Aware Denoising uses first-step temporal variation to reserve full refinement for difficult chunks. Quantitative and qualitative evidence shows that selective memory improves long-range consistency and that adaptive scheduling recovers computation with a controllable quality trade-off. The results motivate future systems in which memory, attention, and iterative refinement are jointly budgeted throughout generation.

\bibliographystyle{abbrv}
\bibliography{references}
\end{document}